\pdfoutput=1
\documentclass[11pt]{article}

\usepackage{acl}

\usepackage{times}
\usepackage{latexsym}
\usepackage{amsmath, amssymb}
\usepackage{dsfont}
\usepackage{tabularx}
\usepackage{graphicx}
\usepackage{booktabs}
\usepackage{makecell}
\usepackage{soul}
\usepackage{svg}
\usepackage{multirow}
\usepackage{tikz}
\usetikzlibrary{positioning, arrows.meta}

\usepackage[T1]{fontenc}
\usepackage[utf8]{inputenc}

\usepackage{microtype}

\title{Zero-shot Cross-Lingual Transfer \\for Synthetic Data Generation in Grammatical Error Detection}

\author{Gaetan Lopez Latouche \and Marc-André Carbonneau \and Ben Swanson \\
        Ubisoft La Forge \\
        \{gaetan.lopez-latouche,marc-andre.carbonneau2,ben.swanson2\}@ubisoft.com}

\begin{document}
\maketitle
\begin{abstract}

%Grammatical Error Detection (GED) methods rely heavily on human annotations from Grammatical Error Correction corpora. 
%\ma{Grammatical Error Detection (GED) methods rely heavily on human annotated Grammatical Error Correction corpora.}
Grammatical Error Detection (GED) methods rely heavily on human annotated error corpora.
However, these annotations are unavailable in many low-resource languages. 
In this paper, we investigate GED in this context. Leveraging the zero-shot cross-lingual transfer capabilities of multilingual pre-trained language models, we train a model using data from a diverse set of languages to generate synthetic errors in other languages. These synthetic error corpora are then used to train a GED model. Specifically we propose a two-stage fine-tuning pipeline where the GED model is first fine-tuned on multilingual synthetic data from target languages followed by fine-tuning on human-annotated GED corpora from source languages. 
This approach outperforms current state-of-the-art annotation-free GED methods. 
We also analyse the errors produced by our method and other strong baselines, finding that our approach produces errors that are more diverse and more similar to human errors. 

% We leverage the cross-lingual transfer capabilities of multilingual pre-trained language models to generate artificial errors in low-resource languages. We train a model using data from a diverse set of languages to generate synthetic errors in other languages, which are then used to train a GED model. 
% Leveraging zero-shot cross-lingual transfer, we train a generative multilingual pre-trained language model using multilingual data to produce synthetic errors in other languages. These synthetic error corpora are then used for to train a GED model.
%However, such annotations are unavailable for many languages preventing the use of supervised approaches.
%In this paper, we investigate zero-shot cross-lingual transfer to generate artificial errors for GED in low-resource languages. 
%
%In addition, we analyse the errors produced by our method and other strong baselines, finding that our approach produces errors that are more diverse and more similar to human errors. 

\end{abstract}

\section{Introduction}

Grammatical Error Detection (GED) refers to the automated process of detecting errors in text. It is often framed as a binary sequence labeling task where each token is classified as either correct or erroneous \cite{volodina-etal-2023-multiged, kasewa-etal-2018-wronging}. GED is widely used in language learning applications and contributes to the performance of grammatical error correction (GEC) systems \cite{yuan-etal-2021-multi, zhou-etal-2023-improving-seq2seq, sutter2024multilingual}. 

Prior research in multilingual GED has primarily operated in supervised settings \cite{volodina-etal-2023-multiged, colla-etal-2023-elicode, yuan-etal-2021-multi}, relying on human annotated data for training. 
Despite recent efforts to obtain annotated corpora \cite{naplava-etal-2022-czech, alhafni-etal-2023-advancements} many languages still lack these resources, motivating research on methods operating without GED annotations. 

To overcome the absence of human annotations, researchers have explored two primary approaches. The first involves language-agnostic artificial error generation (AEG). This is achieved using rules \cite{rothe-etal-2021-simple, grundkiewicz-junczys-dowmunt-2019-minimally}, non-autoregressive translation \cite{sun2022unified}, or round-trip translation \cite{lichtarge-etal-2019-corpora}. These methods are not trained to replicate human errors and compare unfavorably to supervised techniques like back-translation \cite{kasewa-etal-2018-wronging, stahlberg-kumar-2021-synthetic, kiyono-etal-2019-empirical, luhtaru2024err} which train models to learn to generate human errors.

The second approach leverages the cross-lingual transfer (CLT) capabilities of BERT-like \cite{devlin-etal-2019-bert} multilingual pre-trained language models (mPLMs). 
This involves fine-tuning a GED model on languages with abundant human annotations (termed as source languages) and evaluating their performance on languages devoid of human annotations (referred to as target languages). While certain languages exhibit unique error types, most adhere to shared linguistic rules, which mPLMs can exploit to detect errors across languages.

% In this paper, we posit that the generation of errors also share linguistic similarities and introduce a novel approach to leveraging zero-shot CLT. Unlike previous studies that directly leverage zero-shot CLT to the end task, we leverage it for artificial error generation (AEG) across multiple languages through back-translation. Additionally, we propose a two-stage fine-tuning pipeline where a GED model is first fine-tuned on multilingual synthetic data from target languages followed by fine-tuning on human-annotated GED corpora from the source languages to detect errors in any target language.

In this paper, we hypothesize that error generation also share linguistic similarities across languages. We propose a novel approach to zero-shot CLT in GED by combining back-translation with the CLT capabilities of mPLMs to perform AEG in various target languages. Our methodology involves a two-stage fine-tuning pipeline: first, a GED model is fine-tuned on multilingual synthetic data produced by our language-agnostic back-translation approach; second, the model undergoes further fine-tuning on human-annotated GED corpora from the source languages.

% Specifically, we explore a language-agnostic variant of back-translation. 

% \ma{In this paper, we hypothesize that error generation also share linguistic similarities across languages. We propose a novel AEG approach by combining language-agnostic back-translation with the CLT capabilities of mPLMs various languages. Our methodology involves a two-stage fine-tuning pipeline: first, a GED model is fine-tuned on multilingual synthetic data produced by our back-translation approach; second, the model undergoes further fine-tuning on human-annotated GED corpora from the source languages.}

% In this paper, we introduce a novel approach to leveraging zero-shot CLT that combines the zero-shot CLT capabilities of mPLMs to the generation of artificial errors. Unlike previous studies that directly leverage zero-shot CLT to the end task, we leverage it for artificial error generation (AEG) across multiple languages through back-translation. Similarly, to the detection of errors, we posit that the generation of errors share different linguistic rules that a model can exploit to generate errors in target languages. Additionally, we propose a two-stage fine-tuning pipeline where the model is first fine-tuned on multilingual synthetic data from target languages followed by fine-tuning on human-annotated GED corpora from the source languages.  

We experiment on 6 source and 5 target languages and show that our technique surpasses previous state-of-the-art annotation-free GED methods. In addition, we provide a detailed error analysis comparing several AEG methods to ours. 

%Our experiments involving 6 source and 5 target languages show that our technique surpasses previous state-of-the-art annotation-free GED methods. In addition, we provide a detailed error analysis comparing several AEG methods to ours. 

The contributions of this paper are as follows:
\begin{itemize}
    \item We introduce a novel state-of-the-art method for GED on languages without annotations.
    \item We show that we can leverage the CLT capabilities of mPLMs for synthetic data generation to improve performance on a different downstream task, in our case GED.
    \item We provide the first evaluation of GEC annotation-free synthetic data generation methods applied to multilingual GED.
    \item We release a synthetic GED corpus comprising over 5 million samples in 11 languages.
\end{itemize}

\section{Related Work}

\noindent\textbf{GED} Originally addressed through statistical \cite{gamon2011high} and neural models \cite{rei-yannakoudakis-2016-compositional}, GED is now tackled using pre-trained language models \cite{kaneko2019multi, bell-etal-2019-context, yuan-etal-2021-multi, colla-etal-2023-elicode, le-hong-etal-2023-two}.

Historically, most research in GED has been concentrated on the English language. However, recently, \citet{volodina-etal-2023-multiged} organised the first shared task on multilingual GED in which \citet{colla-etal-2023-elicode} set state-of-the-art in all non-English datasets by fine-tuning a XLM-RoBERTa large model on human annotated data in a monolingual setting. While we follow their methodology to train our GED model, we complement prior research by exploring GED for languages lacking annotations.

\noindent\textbf{Artificial Error Generation} Current methods for AEG can be broadly categorized into language-agnostic and language-specific approaches. Language-specific methods focus on replicating the error patterns found in a specific GEC corpora. This can involve heuristic approaches tailored to mimic the linguistic errors identified in GEC corpora \cite{awasthi-etal-2019-parallel,cao-etal-2023-mitigating,naplava-etal-2022-czech}, or employing techniques such as back-translation \cite{kasewa-etal-2018-wronging, stahlberg-kumar-2021-synthetic, kiyono-etal-2019-empirical, luhtaru2024err}. While effective for languages with annotated corpora, these methods are not suitable for languages lacking such resources.

In contrast, there are few language-agnostic methods for generating artificial errors. \citet{grundkiewicz-junczys-dowmunt-2019-minimally} introduce errors in a corpus by deleting, swapping, inserting and replacing words and characters. Replacements rely on confusion sets obtained from an inverted spellchecker. \citet{lichtarge-etal-2019-corpora} introduce noise via round-trip translation using a bridge language. Finally, \citet{sun2022unified} corrupt sentences by performing non-autoregressive translation using a pre-trained cross-lingual language model. All these error generation techniques have primarily been applied to GEC, and to the best of our knowledge, their performance has not been evaluated on GED.

Our work advances existing synthetic data generation methods by exploring a language-agnostic variant of back-translation.

\noindent\textbf{Unsupervised GEC} Unlike GED, GEC without human annotations has been explored in several studies \cite{alikaniotis-raheja-2019-unreasonable, yasunaga-etal-2021-lm, cao-etal-2023-unsupervised}. State-of-the-art unsupervised GEC systems \cite{yasunaga-etal-2021-lm, cao-etal-2023-unsupervised} typically begin with the development of a GED model trained on erroneous sentences generated through rule-based methods \cite{awasthi-etal-2019-parallel} or masked language models \cite{cao-etal-2023-unsupervised}. This GED model is subsequently used with the Break-It-Fix-It (BIFI) method to create an unsupervised GEC system. 

However, the methods used by \citet{yasunaga-etal-2021-lm, cao-etal-2023-unsupervised} for creating the GED model are not language-agnostic, as they rely on a thorough analysis of language-specific error patterns, making them difficult to apply to languages lacking such annotations.

\begin{figure*}
    \centering
    %\includesvg[width=\textwidth]{images/GED_TOP.svg}
    \includegraphics[width=\textwidth]{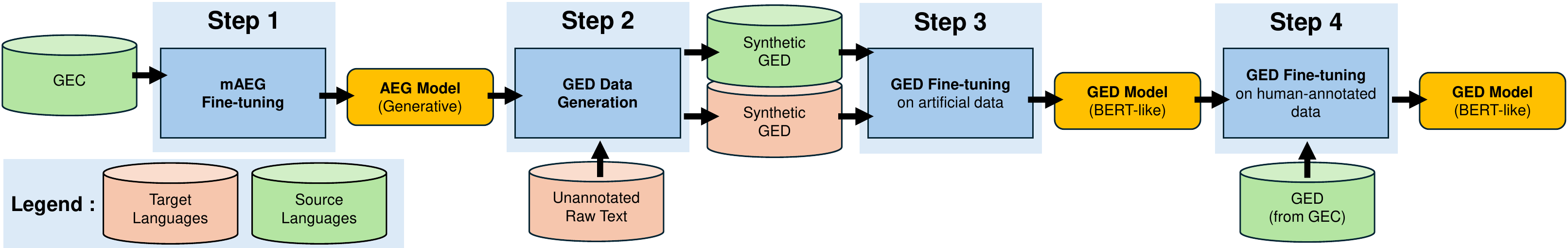}
    \caption{Overview of our proposed method.}
    \label{fig:method_overview}
\end{figure*}

\noindent\textbf{Cross-lingual transfer} Previous studies have shown the capacity of mPLMs to generalize to languages unseen during fine-tuning for both NLU \cite{conneau-etal-2020-unsupervised, chi-etal-2021-infoxlm, Lopez2024align} and generative tasks \cite{xue-etal-2021-mt5, chirkova2024key, shaham2024multilingual}. Close to our work, \citet{yamashita-etal-2020-cross} explored cross-lingual transfer in GEC, a closely related topic. Their findings indicate that pre-training with Masked Language Modeling and Translation Language Modeling enhances cross-lingual transfer. Additionally, they show that fine-tuning on a combination of a high and a low-resource language improves the performance of GEC models on the low-resource language.

In contrast to \citet{yamashita-etal-2020-cross} our research focuses on zero-shot cross-lingual transfer, specifically for GED and AEG, without relying on target language annotations. Additionally, we advance previous work on zero-shot cross-lingual transfer by demonstrating its effectiveness in improving downstream task performance. Investigating zero-shot CLT in GED is particularly significant because the "translate-train" baseline \cite{conneau-etal-2018-xnli, wu2024reuse}, which involves training a GED model on a translated dataset, is infeasible. This arises because machine translation systems tend to correct the errors that the GED model is intended to detect.

\section{Method}\label{method}

Our proposed GED method is developed through a four-step process, as illustrated in Figure \ref{fig:method_overview}. Initially, we train a multilingual AEG model using GEC datasets from the source languages. This AEG model is subsequently employed to produce a GED dataset encompassing both target and source languages. In the third step, we fine-tune a GED model on this multilingual artificially generated dataset. Finally, we perform an additional fine-tuning of the GED model using human-annotated GED data from the source languages. The resultant GED model is capable of detecting errors across any target language.

\noindent\textbf{Data} Our method necessitates three types of corpora. First, the AEG model is trained using GEC datasets in a collection of source languages, $D_s$, which include pairs of ungrammatical sentences and their corrected versions. Additionally, monolingual corpora in the source languages $\Tilde{D}_s$ and in the target low-resource languages $\Tilde{D}_t$, consisting of raw sentences, are required.

\noindent\textbf{AEG Training} The AEG is a generative mPLM trained on a dataset $D_s$ combining all source languages, using the corrected text as input and the ungrammatical one as output. Post-training, the AEG can introduce errors in any language supported by the mPLM, leveraging the inherent zero-shot cross-lingual transfer capabilities of generative mPLMs.

\noindent\textbf{GED Artificial Data Creation} Using our AEG system we obtain a multilingual dataset $D_{synth}$ of raw sentences and their corresponding synthetically generated ungrammatical versions by corrupting sentences from $\Tilde{D}_s$ and $\Tilde{D}_t$. We obtain GED token-level annotation from $D_{synth}$ by tokenizing using language-specific tokenizers, and aligning both sentence versions using Levenshtein distance with minimal alignment following \citet{kasewa-etal-2018-wronging}. We follow the labeling methodology of \citet{volodina-etal-2023-multiged,kasewa-etal-2018-wronging}. We designate tokens that are not aligned with themselves or tokens following a gap as incorrect, while remaining tokens are labeled as correct.

\noindent\textbf{GED model fine-tuning} We propose a two-stage methodology for our multilingual GED model akin to supervised GEC \cite{grundkiewicz-etal-2019-neural, rothe-etal-2021-simple, luhtaru-etal-2024-error}. Models are initially fine-tuned on synthetic data and later refined with human-annotated data. Our approach begins with the fine-tuning of an mPLM such as XLM-R \cite{conneau-etal-2020-unsupervised} on our synthetically generated multilingual GED datasets. Then, we fine-tune this model using human-annotated GED data from all our source languages, $D_s$. 

\begin{table*}
\footnotesize
\centering
\begin{tabular}{llccccc}
    \toprule
     & & \multicolumn{5}{c}{$F_{0.5}$(\%)} \\
    \cmidrule(lr){3-7}
    \textbf{Type} & \textbf{Method} & Swedish & Italian & Czech & Arabic & Chinese \\
    \midrule
    \multirow{3}{*}{\shortstack[l]{Supervised}} & \textsc{\citet{colla-etal-2023-elicode}} & 78.2 & 82.2 & 73.4 & - & -  \\
    & \textsc{\citet{alhafni-etal-2023-advancements}} & - & - & - & 86.6 & -  \\
    & \textsc{\citet{li-etal-2023-templategec}} & - & - & - & - & 59.7  \\
    \midrule
    \multirow{4}{*}{Synthetic data} & \textsc{Rules} & 65.3 & 60.0 & 56.1 & 51.9 & -  \\
    & \textsc{RT translation} & 57.0 & 43.0 & 45.9 & 38.3 & 20.1  \\
    & \textsc{NAT} & 65.9 & 58.6 & 61.1 & 52.5 & 30.4  \\
    % & \textsc{Ours monolingual} & \textbf{70.4} & \textbf{70.3} & \textbf{63.0} & \textbf{62.3} & \textbf{39.8}  \\
    \midrule
    \multirow{2}{*}{Zero-shot} & \textsc{DirectCLT} & 71.5 & 63.8 & 62.1 & 57.3 & 36.2 \\
    & \textsc{Ours} & \textbf{74.7} & \textbf{70.4} & \textbf{66.6} & \textbf{62.8} & \textbf{42.9}  \\
    \bottomrule
\end{tabular}
\caption{Comparison of $F_{0.5}$ between our proposed method, previous synthetic data generation techniques, and the zero-shot cross-lingual transfer baseline on L2 corpora.}
\label{table:main_results}
\end{table*}

\section{Experimental Setup}

\subsection{Datasets \& Evaluation Metric}\label{setup:dataset}

We use English, German, Estonian, Russian, Icelandic, and Spanish as our source languages and Swedish, Italian, Czech, Arabic, and Chinese as our target languages. For each dataset, when multiple subsets are available we use the L2 learners' corpora and the annotations for minimal corrections for grammaticality.

\noindent\textbf{Training set} The English, German, Estonian, Russian, Icelandic, and Spanish datasets are taken from the FCE corpus \cite{yannakoudakis-etal-2011-new}, the Falko-MERLIN GEC corpus \cite{boyd-2018-using}, UT-L2 GEC \cite{rummo2017tu}, RULEC-GEC \cite{rozovskaya-roth-2019-grammar}, the Icelandic language learners section of the Icelandic Error Corpus \cite{arnardottir2021creating}, and COWS-L2H \cite{davidson-etal-2020-developing-nlp}, respectively. We use the training set of each of these GEC datasets to train our generative mPLM. Additionally, for the second stage of our multilingual two-stage fine-tuning pipeline, we use the GED version of each GEC training dataset. For English and German, we use the GED dataset of \citet{volodina-etal-2023-multiged}. For Russian, we convert the $M^2$ files \cite{dahlmeier-ng-2012-better} to a GED dataset following the approach used by \citet{volodina-etal-2023-multiged}; for the remaining languages, we obtain GED annotations from GEC corpora as detailed in \ref{method}.

\noindent\textbf{Evaluation set} The Swedish, Italian and Czech datasets originate from the Swell corpus \cite{volodina2019swell}, MERLIN \cite{boyd-etal-2014-merlin} and GECCC \cite{naplava-etal-2022-czech} respectively. We employ the processed version of those datasets provided in the Multi-GED Shared task 2023 \cite{volodina-etal-2023-multiged}. For Arabic, we use both development and test data of the QALB-2015 shared tasks \cite{rozovskaya-etal-2015-second} provided by \citet{alhafni-etal-2023-advancements}. Finally, the Chinese GED data is derived from two GEC corpora: MuCGEC-Dev \cite{zhang-etal-2022-mucgec} as development set and NLPCC18-Test \cite{zhao2018overview} as test set. We apply the post-processing method described in \ref{method} to produce the GED versions.

\noindent\textbf{Monolingual corpora} Our monolingual text data comes from the CC100 dataset \cite{conneau-etal-2020-unsupervised} in which we sample 200 thousand error-free instances for each language.

\noindent\textbf{Evaluation Metric} Following previous work in GED, we report the token-based $F_{0.5}$ \cite{kaneko2019multi,yuan-etal-2021-multi,volodina-etal-2023-multiged}. For finer-grained analysis we also report the precision-recall curves of our main experiments.

\subsection{Baselines}

We evaluate the proposed artificial error generation method against strong baselines that do not require human-annotated datasets in the target language. We chose methods representative of different family of artificial error generation in GEC: Rules \cite{grundkiewicz-junczys-dowmunt-2019-minimally}, Round-trip translation (RT translation) \cite{lichtarge-etal-2019-corpora}, Non auto-regressive translation (NAT) \cite{sun2022unified}. Additionally, we compare our approach with a zero-shot CLT baseline, which involves directly fine-tuning the GED model on GED datasets from all source languages. We refer to this technique as DirectCLT to distinguish it from our method, which uses the cross-lingual transfer capabilities of generative mPLMs to generate errors in any target language. More information on the implementations of our baselines in Appendix \ref{appendix:baseline_details}.

\begin{table*}
\footnotesize
\centering
\begin{tabular}{lccccc}
    \toprule
    & \multicolumn{5}{c}{$F_{0.5}$(\%)} \\
    \cmidrule(lr){2-6}
    \textbf{Method} & Swedish & Italian & Czech & Arabic & Chinese \\
    \midrule
    \textsc{DirectCLT} & 71.5 & 63.8 & 62.1 & 57.3 & 36.2  \\
    \midrule
     \textsc{Rules} & 65.3 & 60.0 & 56.1 & 51.9 & -  \\
     \textsc{RT translation} & 57.0 & 43.0 & 45.9 & 38.3 & 20.1  \\
     \textsc{NAT} & 65.9 & 58.6 & 61.1 & 52.5 & 30.4  \\
     \textsc{Ours monolingual} & \textbf{70.4} & \textbf{70.3} & \textbf{63.0} & \textbf{62.3} & \textbf{39.8}  \\
     \bottomrule
\end{tabular}
\caption{Comparison of $F_{0.5}$ between the monolingual version of our method and previous synthetic data generation techniques on L2 corpora.}
\label{table:main_results_monolingual}
\end{table*}

\subsection{Models and Fine-tuning setups}\label{setup}

\noindent\textbf{Synthetic Data Generation} We use the No Language Left Behind (NLLB-200) model \cite{nllb2022no} which supports 202 languages as our generative mPLM. Specifically, we use NLLB 1.3B-distilled for all our experiments. Following \citet{luhtaru2024err}, we train the model on non-tokenized text or detokenized if the non tokenized format is not available. Details regarding our hyperparameters can be found in Appendix \ref{appendix:training_details}.

\noindent\textbf{Grammatical Error Detection} In line with \cite{colla-etal-2023-elicode}, we use XLM-RoBERTa-large, a multilingual pre-trained encoder with strong cross-lingual abilities \cite{conneau-etal-2020-unsupervised} as our GED model. We evaluate two versions of our method: (1) A Monolingual version, where the GED model is exclusively trained on synthetic data from the target language, enabling direct comparison with existing synthetic data generation techniques. (2) A Multilingual version using our two-stage fine-tuning procedure to compare against DirectCLT. 

\noindent\textbf{Postprocessing} The postprocessing steps outlined in \ref{method}, which transform synthetic corpora into GED corpora, necessitate tokenized text. To achieve this, we use Stanza \cite{qi2020stanza} for Czech and Spacy \cite{Honnibal_spaCy_Industrial-strength_Natural_2020} for Swedish and Italian. Following previous works on Arabic GEC \cite{belkebir-habash-2021-automatic,alhafni-etal-2023-advancements}, we use CAMeL Tools \cite{obeid-etal-2020-camel}. Lastly, for Chinese, we use the PKUNLP word segmentation tool provided in the NLPCC 2018 shared task \cite{zhao2018overview}.

\begin{figure*}
    \centering
    \includesvg[width=\textwidth]{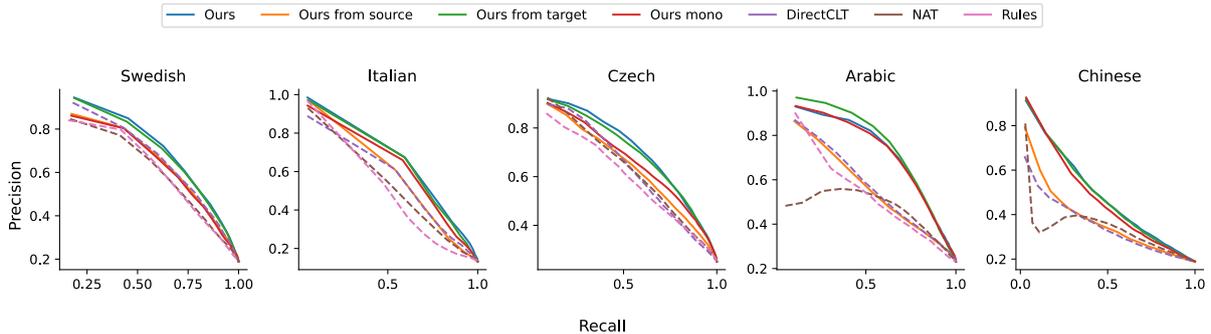}
    \caption{Precision-Recall curves comparing our method in different data configurations to our baselines.}
    \label{fig:pr_curves}
\end{figure*}

\begin{table*}
\small
\centering
\begin{tabular}{lllllll}
    \toprule
    & & \multicolumn{4}{c}{$F_{0.5}$(\%)} \\
    \cmidrule(lr){2-6}
    \textbf{Configuration} & Swedish & Italian & Czech & Arabic & Chinese \\
    \midrule
    \textsc{DirectCLT} & 71.5 & 63.8 & 62.1 & 57.3 & 36.2  \\
    \textsc{Ours} & \textbf{74.7} & 70.4 & 66.6 & 62.8 & 42.9\\
    \textsc{Ours from source} & 72.5 & 64.1 & 62.9 & 58.4 & 36.5  \\
    \textsc{Ours from target} & 74.2 & \textbf{71.3} & \textbf{67.3} & \textbf{71.6} & \textbf{47.9}  \\
    % \textsc{Ours opposite} \\
    % \textsc{\MyIndent to source} & 70.6 & 65.0 & 60.6 & 58.6 & 31.1  \\
    % \textsc{\MyIndent to target} & 72.1 & 70.4 & 63.4 & 60.7 & 39.5  \\
    % \textsc{\MyIndent to source and target} & 72.2 & 69.2 & 63.1 & 59.3 & 39.9  \\
    % \textsc{Ours monolingual} & \textbf{70.4} & \textbf{70.3} & \textbf{63.0} & \textbf{62.3} & \textbf{39.8}  \\
    \bottomrule
\end{tabular}
 \caption{Comparison of $F_{0.5}$ of our method where first-stage fine-tuning is performed on various data configurations.}
\label{table:ged_multilingual_impact}
\end{table*}

\section{Proposed Method Evaluation}

\subsection{Comparison to State-of-the-Art}

Table \ref{table:main_results} presents the performance of our method compared to previous state-of-the-art. Our method establishes a new standard in GED without human annotations across all target languages, outperforming both synthetic data generation techniques and DirectCLT by a significant margin.

We posit that our superior performance can be attributed to the capability of our AEG method to produce a diverse set of errors including language-specific errors. This hypothesis is further examined in Section \ref{analysis}.

It is worth mentioning that while our results represent a significant advancement, they still fall short of the state-of-the-art supervised settings. This result is expected and aligns with the existing literature in GED, which highlights notable discrepancies when evaluating supervised models with out-of-domain data, even if it originates from the same language as the training data \cite{volodina-etal-2023-multiged, colla-etal-2023-elicode}.

\subsection{Evaluation of AEG}

As all previous work using AEG for GED has been in monolingual settings, we introduce a monolingual variant of our approach. Here, the GED model is exclusively fine-tuned on synthetic data from the target language. 

Table \ref{table:main_results_monolingual} shows that our synthetic data generation technique achieves the best performance among annotation-free synthetic data generation methods applied to GED. Given that rule-based methods apply a set of transformations without considering the sentence context, the average improvement of 9.2 points of $F_{0.5}$ over these methods highlights the significance of generating context-dependent errors in synthetic data generation. Additionally, given that NAT is not trained to generate errors but to produce translations, outperforming this method by 8.3 points of $F_{0.5}$ highlights the advantage of learning to generate errors from authentic instances, even when these instances originate from different languages.

We hypothesize that the ability to synthesize context-dependent errors combined with the acquisition of error-generation insights from authentic instances empower our method to yield errors more akin to human errors, thus leading to better performance. We further analyze this hypothesis in \ref{case_study}.

Additionally, our monolingual setup outperforms DirectCLT in four out of five languages. This is a notable achievement given other synthetic data generation methods' inability to meet this benchmark. Both approaches leverage the CLT of mPLMs, albeit differently: ours uses it for artificial error generation in target languages with a generative mPLM, while DirectCLT leverages it directly to perform error detection across target languages. This comparison suggests that our method creates tailored error patterns in target languages that a GED model trained only on source language annotations cannot detect, indicating that our approach to CLT in GED could generalize to other NLU tasks, which is a promising avenue for future research.

\subsection{Language Ablation}

We study the effect of changing the language configuration of the synthetic data. We compare fine-tuning the GED model using synthetic data comprising different language sets: exclusively source languages, exclusively target languages, and a combination of both source and target languages.

Results in Table \ref{table:ged_multilingual_impact} show that any first stage fine-tuning language configuration improves the GED performance of our method over the DirectCLT baseline, highlighting the robustness of our two-stage fine-tuning pipeline. Notably, including synthetic data from the target language results in a more significant improvement which emphasize the importance of using a language-agnostic artificial error generation method capable of generating errors in any target language. 

Furthermore, results from Table \ref{table:ged_multilingual_impact} suggest that first-stage fine-tuning exclusively on synthetic data from target languages outperforms fine-tuning on a combination of source and target languages. However, comparing $F_{0.5}$ scores does not reveal the big picture and can lead to false conclusion. The $F_{0.5}$ score is computed at an operation point that is usually arbitrarily set to 0.5 in the literature \cite{kasewa-etal-2018-wronging, colla-etal-2023-elicode, le-hong-etal-2023-two}. For a more comprehensive comparison of performance, Figure \ref{fig:pr_curves} presents the Precision-Recall curves for each method. It shows that fine-tuning on either synthetic data from source and target languages or target languages alone yields similar results. We can conclude that the determining factor is the inclusion of synthetic data in the target language. We can also see that our method outperforms other baseline in the curves too. We encourage practitioners to use such figures to compare GED models for more meaningful conclusions than threshold dependant metrics such as $F$ scores.

We experimented with reversing our fine-tuning pipeline by initially training on human annotations from our source languages followed by fine-tuning on synthetic data. However, this approach empirically yielded inferior performance. The fact that ending the fine-tuning process with human-annotated data, even in source languages, is more effective than using target language synthetic data indicates that artificial errors still do not reach the quality of authentic corpora. Otherwise it would make sense to end the training with errors specific to the target language. We hypothesize that improved synthetic error generation techniques would lead to opposite conclusions regarding the fine-tuning order. 

% Intuitively, having source languages from the same family as the target language should improve transfer quality. We experimented with languages from the Slavic and Romance language families. We measured the impact of removing a source language belonging to the same family against the removal of a source language from an unrelated language family from the fine-tuning dataset of our AEG. On our limited set of possible experiments we did not find large difference in performance which is in line with other zero-shot CLT findings \cite{shaham2024multilingual, wu2024reuse}.

\begin{figure}
    \centering
    \includesvg[width=\columnwidth]{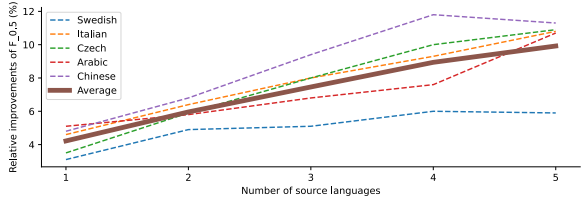}
    \caption{Relative improvement in terms of $F_{0.5}$ score compared to English-only fine-tuning as additional source languages are incorporated.}
    \label{fig:nb_lang}
\end{figure}

\subsection{Scalability}\label{results:scalability}

\begin{table}
\small
\centering
\begin{tabular}{llll}
    \toprule
    & Czech $L1$ & Arabic $L1$ \\
    \midrule
    {RT translation} & 20.2 & 38.7  \\
    {Rules} & 26.5 & 32.9  \\
    {NAT} & 38.0 & 48.9 \\
    {DirectCLT} & 41.7 & 45.5 \\
    {Ours} & \textbf{41.8} & \textbf{63.2} \\
    \bottomrule
\end{tabular}
 \caption{$F_{0.5}$ (\%) on out-of-domain L1 corpora.}
\label{table:cross_domain}
\end{table}

Here we investigate how our synthetic data generation method scales as new languages corpora become available. We fine-tune the AEG model by progressively incorporating new languages in different orders to an English-only fine-tuned baseline. We follow the protocol of \citet{shaham2024multilingual}. We report average scores per target language of a GED model fine-tuned on monolingual synthetic data.

Figure \ref{fig:nb_lang} shows that on average, performance increases with the number of source languages. This suggests that our synthetic data generation method applied to GED might continue to improve as new GED corpora become available. 

\subsection{Generalization to out-of-domain errors}

Errors vary between different populations. For instance native speakers (L1) do not commit the same type of errors than second language learners (L2). We investigate the robustness of our method to different error distributions. Our method is trained on L2 learner corpora and we evaluate it on L1 data. We found available GED annotated data of L1 speakers for Arabic and Czech:  QALB 2014 \cite{mohit-etal-2014-first} and the Native Formal section of GECCC \cite{naplava-etal-2022-czech}.

Table \ref{table:cross_domain} presents the results. Our method surpasses all other baselines, demonstrating its continued suitability for out-of-domain corpora in the target language. Unlike the other baselines, our method achieves approximately similar performance on both L1 and L2 Arabic corpora. However, for Czech, all methods show a significant decrease in performance. 
We hypothesize that this is due to the unique stringent rules regarding the use of commas in Czech. This results in the predominance of "Punctuation" errors in the L1 Czech corpora, which are less common in many other languages, and therefore amplify the difference between domains.

\section{Analysis of synthetic errors}\label{analysis}

We compare the errors produced by the AEG methods. We first study Czech using a Czech extension \cite{naplava-etal-2022-czech} of the ERRANT \cite{bryant-etal-2017-automatic} error annotation tool and an artificial vs human error discriminator. We then extend our analysis to many languages using GPT-4 \cite{openai2024gpt4} to classify error types.

\begin{table}
\small
\centering
\begin{tabular}{lllll}
    \toprule
     & Precision & Recall & $F_1$    \\
    \midrule
    {Rules} & 96.5 & 95.2 & 96.6 \\
    {NAT} & 94.3 & 97.2 & 95.2  \\
    {Ours} & 79.1 & 88.3 & 83.4 \\
    \bottomrule
\end{tabular}
 \caption{Performance of a binary classifier trained to distinguish between human errors and errors produced by a synthetic data generation technique. We report the Precision, Recall and $F_1$ score.}
\label{table:fake_vs_real_errors}
\end{table}

\subsection{Czech Case Study}\label{case_study}

\noindent\textbf{Similarity Analysis with Human Errors} To assess if the synthetic instances are realistic and human-like, we train a binary classifier (one per synthetic data generation technique) to distinguish between errors generated by a particular synthetic data generation method and human errors. We constructed a development set comprising approximately equal numbers of authentic and synthetic data and assessed performance using the $F_1$ score.  More information on how we train the classifier can be found in \ref{appendix:classifier_analysis}. Results are presented in Table \ref{table:fake_vs_real_errors}.

Our classifier achieves an $F_1$ score of 83.4\% for the proposed method, indicating a moderate ability to differentiate between synthetic and human errors. This supports our hypothesis that our synthetic data generation method does not fully replicate the quality of authentic sentences. In contrast, the classifier achieves an $F_1$ score exceeding 95\% for other synthetic data generation methods, suggesting a higher degree of differentiation. Overall, this suggests that our method produces errors that are more human-like, translating into better downstream performance.

\begin{figure}
    \centering
    \includesvg[width=\columnwidth]{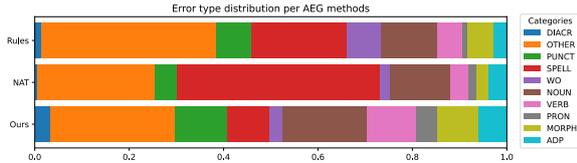}
    \caption{Top 10 error types distribution of different annotation-free synthetic data generation methods.}
    \label{fig:czech_errant}
\end{figure}

\noindent\textbf{Error Distribution} We use the Czech extension \cite{naplava-etal-2022-czech} of ERRANT to categorize the errors made by different systems. Figure \ref{fig:czech_errant} presents the distribution of the top 10 error types for the various synthetic data generation methods studied. Our method produces a more diverse set of errors compared to NAT \cite{sun2022unified} and rule-based approaches \cite{grundkiewicz-junczys-dowmunt-2019-minimally}. Notably, while other methods predominantly yield 'Other' and 'Spell' error types, our method features a more balanced distribution of error types, indicating that our method is more effective in mimicking the complexity and range of human language errors. 

Additionally, our method generates a higher percentage of 'DIACR' errors compared to other techniques. Since 'DIACR' errors are the most common among L2 learners of Czech, this could explain the performance improvements of our method.
Given that 'DIACR' errors are specific to Czech \cite{naplava-etal-2022-czech} in the set of languages we study, this indicates that our method can produce error types not encountered during the fine-tuning on source languages of our generative mPLM. 

\begin{figure}
    \centering
    \includesvg[width=\columnwidth]{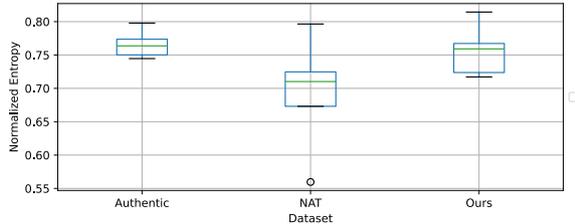}
    \caption{Normalized Entropy comparison of authentic and synthetic errors aggregated over different datasets.}
    \label{fig:gpt_errant}
\end{figure}

\subsection{Multilingual Extension}

We want to extend our previous findings by assessing if our synthetic data generation method effectively captures a variety of error types across all languages. For this, we need a language-agnostic classifier. 
We use GPT-4 to classify errors from various sources across all the languages under investigation. 
Prior studies have shown that GPT-4's judgments align closely with human evaluations \cite{wang-etal-2023-chatgpt, fu2023gptscore} and exhibit promising error correction capabilities \cite{fang2023chatgpt, davis2024prompting, wu2023chatgpt}. Although a thorough assessment of GPT-4 for error classification is beyond the scope of the study, we performed a limited qualitative analysis of GPT-4's accuracy in Italian, Swedish, Spanish, and English with native speakers. We found that it is suitable for our application. For each type of error classified by GPT-4 we compute its frequency distribution across data and compute the entropy of this distribution.  Further details on our evaluation methodology are provided in Appendix \ref{appendix:gpt4_analysis}. 

Figure \ref{fig:gpt_errant} validates our previous findings that our method generates a more diverse set of errors compared to NAT. However, the range of error types generated by our method is narrower than that produced by humans. Moreover, the variability in the diversity of error types is significantly higher with our method than with human errors across different languages. This suggests that our method does not consistently perform across languages.  

\section{Conclusion}

We introduced a novel zero-shot approach for GED with low-resource languages. Our method combines back-translation with the CLT capabilities of mPLMs to perform AEG across various target languages. Then, we fine-tune the GED model in two steps: first on multilingual synthetic data from source and target languages, then on human-annotated source language corpora. This method achieves state-of-the-art performance in annotation-free GED. Our error analysis shows that we produce errors that are more diverse and human-like than the baselines. 

In future work, we intend to explore the potential of our GED models to enhance unsupervised GEC methods.

\section{Limitations}

Our approach relies on the CLT capabilities obtained during the multilingual unsupervised pre-training of mPLMs. Consequently, the applicability of our method is restricted to the languages supported by the mPLM. Furthermore, its performance on each language may vary depending on the amount of pre-training data available for that language. This limitation is inherent to all studies leveraging mPLMs.

Additionally, our study primarily focuses on the errors made by second language learners. While we have analyzed the performance of our method on native language corpora, it would be valuable to evaluate its generalizability to other domains within a language. For instance, this includes errors made in casual text messaging or by machine translation systems.

Compared to the direct application of CLT in GED, our method involves additional steps such as training a generative mPLM and generating a substantial amount of synthetic data. These requirements may pose challenges for researchers with limited computational resources and could limit the practicality of developing this approach in resource-constrained environments. To address this constraint, we have made available a synthetic GED corpus encompassing more than 5 million samples across 11 languages.

\section{Ethics Statement}

Our research is driven by a commitment to supporting and preserving linguistic diversity. Low-resource languages often face marginalization in the realm of technological advancements. By developing GED models for these languages, we aim to enhance their digital presence and usability, thus promoting linguistic equity. 

However, it is important to acknowledge potential ethical concerns. The use of CLT to generate synthetic data, while beneficial for training GED models, carries the risk of misuse. Such systems could potentially be exploited to create false information or propaganda in low-resource languages. Additionally, while GED systems are crucial for regions with a shortage of language teachers, there is a risk that their widespread use could lead to an over-reliance on these tools. This dependency might result in a decline in the linguistic and grammatical skills of native speakers, as they become more reliant on technology for language correction and validation.

It is essential for future users to use these technologies judiciously. Balancing the use of GED tools with a genuine effort to improve one's linguistic abilities is crucial. Building on the research by \citet{fei-etal-2023-enhancing} could provide a valuable advancement by incorporating explainability into our GED systems.

\bibliography{anthology,custom}
\bibliographystyle{acl_natbib}

\clearpage

\appendix

\section{Appendix}
\label{sec:appendix}

\subsection{Baselines}\label{appendix:baseline_details}

\noindent\textbf{Rules} We re-implemented \citet{grundkiewicz-junczys-dowmunt-2019-minimally} using Aspell dictionaries\footnote{http://aspell.net/} for the replacement operation.

\noindent\textbf{NAT} We replicated the NAT model using InfoXLM \cite{chi-etal-2021-infoxlm} and English as source language, following  \cite{sun2022unified} methodology.  For non-autoregressive translation generation, we used Europarl \cite{koehn-2005-europarl} for Italian, Swedish and Czech and the UN Parallel Corpus v1.0 \cite{ziemski-etal-2016-united} for Arabic and Chinese. We conducted hyper-parameter tuning for the NAT-based data construction by exploring the parameter set specified in \cite{sun2022unified} and selected the optimal parameters for each language based on performance on the development set.

\noindent\textbf{RT translation} We use OPUS-MT \cite{tiedemann-thottingal-2020-opus} as our translation model and English as the bridge language. 

\subsection{Implementation details}\label{appendix:training_details}

\noindent\textbf{Artificial error generation} We use two distinct AEG models to generate errors in target and source languages, both based on NLL 1.3B-distilled but trained with different hyper-parameters.

For synthetic data generation in target languages, we conduct preliminary grid searches on the Swedish development set to determine the optimal hyperparameters. We select the learning rate from \{1e-4, 5e-4, 1e-5, 5e-5\} and the number of epochs from \{3, 5, 10, 15, 20\}. Ultimately, we set the learning rate to 1e-5 and fine-tune for 3 epochs with a batch size of 24 and a linear scheduler.

For synthetic data generation in source languages, we use a different set of hyper-parameters based on grid searches on the English development set. The learning rate is set to 1e-4, and we fine-tune for 10 epochs with a batch size of 24 and a linear scheduler.

\noindent\textbf{Grammatical error detection} Based on initial experiments with the Swedish development set, we use a learning rate of 1e-5, a batch size of 24, and train for 5 epochs with a linear scheduler. In our second-stage experiments, we maintain the same setup but fine-tune for only 1 epoch.

\noindent\textbf{Monolingual corpora:} As mentioned in Section \ref{setup:dataset}, our monolingual text data is sourced from the CC100 dataset \cite{conneau-etal-2020-unsupervised}, from which we sample 200,000 error-free instances for each language. To ensure the text is error-free, we use the DirectCLT baseline for error detection, including only sentences verified to be error-free.

For all our trainings, we use 3*A6000 GPUs with 48 GB of VRAM.

\subsection{Similarity Analysis details}\label{appendix:classifier_analysis}

To distinguish between authentic and synthetic instances, we train a binary classifier. The classifier processes a pair of sentences: a grammatical sentence and its corresponding ungrammatical version separated by a separator token. Its task is to identify whether the ungrammatical sentence is synthetic or authentic. We train separate binary classifiers for each synthetic data generation method, using mdeberta-v3-base \cite{he2023debertav3} as our backbone.

\subsection{GPT-4 analysis details}\label{appendix:gpt4_analysis}

To evaluate the linguistic diversity of errors across different languages, we employed GPT-4 as an error classifier. Specifically, we used GPT-4 to describe the nature of the errors in sentences. Without constraining GPT-4 to a predetermined set of error types, it generated a diverse range of error descriptions for similar errors.

We then categorized these errors into distinct clusters using a clustering method based on the sentence embeddings generated using sentence-transformers \cite{reimers-2019-sentence-bert}. In particular, we applied KMeans clustering with four different values of K (16, 32, 64, 128). This approach produced multiple sets of clusters, each representing distinct error patterns within the dataset.

For each value of K, we computed the frequency distribution of errors across the clusters and subsequently calculated the entropy of these distributions. To enable comparison across different values of K, we normalized the entropy values, ensuring comparability and eliminating bias from the number of clusters chosen.

Finally, to derive a comprehensive measure of normalized entropy for each language under study, we averaged the normalized entropy values obtained across all K settings. The resulting normalized entropy metric provides a robust indicator of the diversity of error patterns observed across different languages, as illustrated in Figure \ref{fig:gpt_errant}.

\begin{table}
\small
\centering
\begin{tabular}{lllllll}
    \toprule
    1 & 2 & 3 & 4 & 5 & 6    \\
    \midrule
    en & en,de & en,de,is & en,de,is,et & en,de,is,et,ru & all \\
    en & en,es & en,es,de & en,es,de,et & en,es,de,et,is & all  \\
    en & en,is & en,is,es & en,is,es,ru & en,is,es,ru,de & all \\
    \bottomrule
\end{tabular}
 \caption{Subsets of source languages used to fine-tune our AEG model for
our scalability experiments in \ref{results:scalability}}.
\label{table:impact_of_nb_lang}
\end{table}

\end{document}